\documentclass{article}

\usepackage{PRIMEarxiv}

\usepackage[utf8]{inputenc} 
\usepackage[T1]{fontenc}    
\usepackage{hyperref}       
\usepackage{url}            
\usepackage{booktabs}       
\usepackage{amsfonts}       
\usepackage{nicefrac}       
\usepackage{microtype}      
\usepackage{lipsum}
\usepackage{fancyhdr}       
\usepackage{graphicx}       
\usepackage{amsmath}
\graphicspath{{./figures/}}     
\usepackage{float}
\usepackage{graphicx}
\usepackage{placeins}
\usepackage{tabulary}
\usepackage{tabularx}
\usepackage{appendix}
\usepackage{pdfpages}
\usepackage{subcaption}
\usepackage{placeins}
\usepackage{tcolorbox}

\title{Detecting Crop Burning in India using Satellite Data}

\vspace{-20mm}
\author{ \\
  Kendra Walker,\thanks{Corresponding author, \texttt{kendrawalker@ucsb.edu}. We thank Tamma Carleton, Gabriel Daldegan and Esther Rolf for guidance on the methods, and Maggie Klope, Stefania di Tommaso, Maddie Berger and Derek Nguyen for excellent research assistance. The work was supported by a generous gift from Jody and John Arnhold.} \space Ben Moscona, Kelsey Jack, Seema Jayachandran, \\
  Namrata Kala, Rohini Pande, Jiani Xue, Marshall Burke \\
   \\
  \vspace{-20mm}
}
\begin{document}
\maketitle
\begin{abstract}
 Crop residue burning is a major source of air pollution in many parts of the world, notably South Asia. Policymakers, practitioners and researchers have invested in both measuring impacts and developing interventions to reduce burning. However, measuring the impacts of burning or the effectiveness of interventions to reduce burning requires data on where burning occurred. These data are challenging to collect in the field, both in terms of cost and feasibility. We take advantage of data from ground-based monitoring of crop residue burning in Punjab, India to explore whether burning can be detected more effectively using accessible satellite imagery. Specifically, we used 3m PlanetScope data with high temporal resolution (up to daily) as well as publicly-available Sentinel-2 data with weekly temporal resolution but greater depth of spectral information. Following an analysis of the ability of different spectral bands and burn indices to separate burned and unburned plots individually, we built a Random Forest model with those determined to provide the greatest separability and evaluated model performance with ground-verified data. Our overall model accuracy of 82-percent is favorable given the challenges presented by the measurement. Based on insights from this process, we discuss technical challenges of detecting crop residue burning from satellite imagery as well as challenges to measuring impacts, both of burning and of policy interventions.
\end{abstract}


\section{Introduction}
Crop residue burning has become a major environmental problem that exacerbates global climate change as well as local health problems from exposure to smoke and particulate matter. In northern India, residue burning of rice stubble left behind after harvest contributes to a cloud of pollution in early winter months \cite{Bhuvaneshawari2019, Mor2021, Sahu2021, Sharma2010} that affects the health and daily lives of local residents \cite{Gupta2020,Lohan2018}. Residue burning has increased in recent decades \cite{Jethva2019,Ravindra2019} due to increased rice production \cite{Jethva2019}, intensified cropping schedules that require a quick turn around between harvesting rice and sowing the winter crop \cite{Liu2021}, and mechanized harvesters that leave rice stalks (stubble) that are too big to be tilled back into the soil \cite{Badarinath2006,Chawala2020}. To mitigate the pollution from burning, alternative strategies and incentive mechanisms have been proposed to help farmers manage crop stubble in a more sustainable manner \cite{Gupta2014}. While assessing the impact of such strategies is essential for successful mitigation, measuring uptake and effectiveness of individual strategies remains a challenge.

Remote sensing offers robust tools to measure burning across landscapes, yet the fine spatial and temporal nature of burning of individual rice paddies puts such monitoring at the edge of current capabilities. Methods to monitor fire via coarse-resolution sensors such as MODIS and AVHRR have been developed over decades to provide active and post-fire measures for all regions including northern India \cite{Vadrevu2011}. While the daily imaging frequency of these satellites allows for near real-time observations, coarse pixel resolution makes it impossible to attribute observations to specific plots, especially in the case of smallholder farming. Sensors specifically designed for India, such as AWiFS and LISS offer finer spatial resolution (56m and 23.5m, respectively) and have been used for more precise localization of fires \cite{Badarinath2006,PRSCL2015,Singh2009}. Landsat and more recently Sentinel-2 imagery have also been used to pinpoint fire locations and estimate burn areas in northern India \cite{Bar2020,Chawala2020,Singh2021,Deshpande2022}. The higher spatial resolution of these sensors comes at the expense of lower temporal resolution, however. In the case of crop residue burning, there is often a very narrow window to observe burning, as a plot may be completely burnt within a day and resemble unburned plots within two to three days. Longer windows between observations might result in underestimation of burn events.

This paper takes advantage of satellite imagery with both high spatial and temporal resolution, recently made available at an unprecedented scale by Planet Labs. The measurement described here was developed as part of a broader study to evaluate the effect of conditional payments on burning, discussed in more detail in Jack, Jayachandran, Kala and Pande \cite{JJKP}. The objective of this paper is to present the data and methods used to model crop residue burning with remote sensing data and to discuss challenges to the modelling. Due to the field work component of the broader study, we have access to ground data verifying the presence or absence of crop burning with which to assess model accuracy; such data is rare for such remote sensing projects and provides insight into the potential and challenges of crop-burn modelling.

\section{Data and Methods}
\label{sec:methods}
\subsection{Ground Data}
\begin{figure}
  \centering
  \includegraphics[width=15cm]{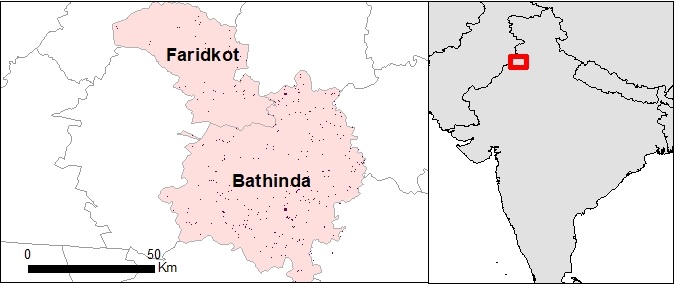}
  \caption{Study plots distributed across study area in Punjab State, India}
  \label{fig:fig1}
\vspace{-3mm}
\end{figure}

Ground-level measures were collected in 2019 as part of a study conducted in two districts in Punjab State, India (Fig \ref{fig:fig1}) and consist of perimeter measures for 3206 rice plots, with burning outcomes (labels) collected for 240 unburned and 441 burned plots. Plots are small, with an average size of 1.4 ha and median of 0.9 ha. All burning occurred between October 2019 and the beginning of the winter planting season in mid-December, although specific burn dates are unknown in most cases. For 321 plots, farmers participating in the study invited a monitor to visit to confirm that the stubble was managed without burning. In 52 of these cases, evidence of burning was found.\footnote{Evidence of burning includes active or observable past burning of straw, root/stem residues, standing stems or other stubble; presence of black/grey ash on the soil surface; or burnt grass, weeds, leaves or branches around the edge of the plot. If any sign of burning was detected, the entire plot was marked as burned.} An additional 23 plots were removed due to signs of burning during an unannounced spot check on a different day. The remaining 246 confirmed unburned plots comprise our preliminary no burn labels. An additional 755 plots received unannounced visits to check for burning. Plots without evidence of burning were marked as unburned at that moment, but not used as labels due to inability to confirm that burning did not occur some time after (or long before) the visit. Due to the random timing of the spot checks, it was not possible to observe stubble management for these plots to confirm investment in an alternative to burning. The 52 plots where burn indicators were noted during invited visits and 401 plots where burn indicators were noted during random spot checks comprise our preliminary set of burned labels. Due to technical issues in recording the perimeter of some plots, 18 labels were removed from our dataset, resulting in a final set of 441 burned and 240 unburned labels.   

\subsection{Satellite Data and Preprocessing}
We use 3m-resolution PlanetScope data as the primary source of imagery for this study. Planet’s large constellation of more than 100 small cubesats in low-Earth orbit provides frequent observations for any location. We downloaded the four-band surface reflectance product from October 10 to December 15, 2019, with a maximum cloud cover of 10 percent. This resulted in an average of 30-40 images per plot.

The PlanetScope Surface Reflectance product includes correction for atmospheric conditions based on the 6SV2.1 radiative transfer code and MODIS near-real-time data inputs \cite{Planet2021}. A recently-released harmonization tool was used to transform Surface Reflectance measurements from the Dove-Classic (PS) and Dove-R (PS2.SD) to approximate measurements at target Sentinel-2 images, helping to normalize the spectral response functions of the different constellations \cite{Kington2022}. The unusable data masks (UDM2) provided with the imagery were applied to remove clouds remaining in the imagery after the 10 percent maximum cloud cover filter.

Sentinel-2 imagery was included to test the additionality of more robust spectral information. All Sentinel-2 images overlapping the study area and period were downloaded from USGS as level-1C products, with geometric and radiometric corrections already applied. The top-of-the atmosphere reflectance was then converted to bottom-of-the-atmosphere (surface) reflectance using the Sentinel-2 toolbox in SNAP.\footnote{A Bidirectional Reflectance Distribution Function (BRDF) model could be applied at this step to correct for variability in illumination, but we chose to skip this due to lack of such correction in the PlanetScope data. The assumption of Lambertian scattering is reasonable given the relatively flat terrain of our study area.} 
Corresponding cloud-probability (S2cloudless) layers were downloaded from Google Earth Engine and pixels with cloud probabilities >=.5 masked as clouds. For the task of detecting clouds, these masks performed better than the data in the native Sentinel-2 SCL layer as well as the QA60 bands available in Google Earth Engine. All three methods, however, missed the majority of shadows. Shadows can be particularly problematic in that they are spectrally similar to burned pixels \cite{Huang2016}. The dark-area data from the SCL layer provided with the Sentinel products covers some of the missed shadows, but are unacceptable for this task, as they confuse recent burn scars with shadows and thus remove most of these as well. The fact that dark shadows closely resemble burn scars underscores the need for careful treatment of shadows in a post-burn detection model. We treated shadows by inspecting each Sentinel image in ESRI ArcGIS and manually expanding masks derived from the s2cloudless products to cover unmasked shadows.

Sentinel-2 resolution was matched to the 3m PlanetScope resolution by simple up-sampling based on cubic convolution. More sophisticated methods have been used to pan-sharpen the Sentinel-2 20m resolution Red-edge and SWIR bands based on the average value of the visual and the near infrared bands \cite{Kaplan2018}, or even to pan-sharpen all Sentinel-2 bands with the 3-m PlanetScope bands \cite{Kaplan2020, Li2020,Sadeh2021}. Given the noted spectral distortion that generally occurs with pan-sharpening, and likelihood that such effect would be augmented in the relatively noisy PlanetScope data, we did not think this was warranted for our data and purposes. 

\vspace{-2mm}
\subsection{Feature Data}
\subsubsection{Spectral Indices}
Even following the calibration steps described above, spectral values of Planet imagery were observed to fluctuate considerably for images with close temporal proximity. Spectral indices were used to help temper this image-to-image fluctuation by normalizing band values with respect to other band values. Most indices commonly used to detect burn scars do not apply to 4-band PlanetScope data due to limitations in spectral resolution. With only Blue, Green, Red and NIR bands to consider, we calculated two standard indices (Simple Ratio and NDVI), two char indices presented in the literature to capture the visible properties of charred soil \cite{Chuvieco2002,Fraser2017}, and one bare soil index \cite{Jamalabad2004} (see Figure \ref{fig:fig2}). Our study compares burnt plots with plots that are harvested and often tilled; our primary separation task is thus between bare soil and charred soil. This is different from many burn-detection tasks, such as those involving wildfire and slash-and-burn events, in that vegetation change is not a key indicator. The lesser-used char index is thus expected to perform better than the more common vegetation-based indices, such as NDVI (and SAVI, GEMI, etc.) for our task.  

While PlanetScope data provides superior temporal resolution and thus higher probability of capturing actual burn events, missed events do occur due to cloud cover or other gaps in observations. Spectral signatures of the visible and NIR bands rapidly become inseparable from unburned plots within days of burning, thus a lack of observations within a day or two of the burn event likely result in misclassification as unburned regardless of index. While providing slightly lower temporal and spatial resolution than PlanetScope, observations from Sentinel-2 data can complement those of PlanetScope by providing information in the mid-infrared range, which is less sensitive to noise from aerosols and haze from biomass burning \cite{Kaufman&Remer1994} and generally better for separating burned and unburned land \cite{Amos2019,Singh2021,Trigg2000}. Additional spectral indices calculated with Sentinel-2 data include two forms of the Normalized Burn Index (NBR, NBR2), used in a wide array of fire-detection tasks; the Mid-Infrared Bispectral Index (MIRBI) \cite{Trigg2001}, which has shown strong performance in distinguishing fires in landscapes as diverse as the Peruvian Amazon \cite{Barboza2020}, Mexican wetlands \cite{Perez2022}, Mediterranean \cite{Smiraglia2020,Damasio2018} and African grassland \cite{Trigg2001,Vhengani2015}; and the Burn Scar Index (BSI), developed specifically for crop-residue burning \cite{Wang2018}. We also derived a char index from Spectral Mixture Analysis based on Burned Area Spectral Mixture Analysis (BASMA) methods developed by Daldegan et al., 2019 \cite{Daldegan2019}.

\begin{figure}
\vspace{-2mm}
\begin{tcolorbox}
\vspace{3mm}
\textbf{Indices derived from PlanetScope imagery}
\begin{equation}
Simple Ratio \hspace{1 mm}\textbf{(SR)} = \hspace{100mm} \frac{NIR}{Red}
\end{equation}
\begin{equation}
Normalized Difference Vegetation Index  \hspace{1 mm}\textbf{(NDVI)} = \hspace{42mm} \frac{NIR-Red}{NIR+Red}
\end{equation}
\begin{equation}
Char Index  \hspace{1 mm}\textbf{(CI)} = (Blue + Green + Red) + Max(|Blue - Green|, |Blue - Red|, |Red - Green|) {*}15
\end{equation}
\begin{equation}
Burn Area Index \hspace{1 mm}\textbf{(BAI)} = \hspace{55mm} \frac{1}{(.06-\frac{NIR}{10000})^2+(.1-\frac{Red}{10000})^2}
\end{equation}
\begin{equation}
Bare Soil Index \hspace{1 mm}\textbf{(BSoI)} = \hspace{35mm}  \frac{(NIR + Green) - (Red + Blue)}{NIR + Green + Red + Blue}{*}100 + 100
\end{equation}
\textbf{Indices derived from Sentinel-2 imagery}
\begin{equation}
Normalized Burn Index  \hspace{1 mm}\textbf{(NBR)} = \hspace{65mm} \frac{NIR-SWIR2}{NIR+SWIR2}
\end{equation}
\begin{equation}
Normalized Burn Index 2  \hspace{1 mm}\textbf{(NBR2)} = \hspace{58mm} \frac{SWIR1-SWIR2}{SWIR1+SWIR2}
\end{equation}
\begin{equation}
MidInfrared Bispectral Index  \hspace{1 mm}\textbf{(MIRBI)} = \hspace{32mm} 10{*}SWIR2 - 9.8{*}SWIR1 + 2
\end{equation}
\begin{equation}
Burn Scar Index  \hspace{1 mm}\textbf{(BSI)} = \hspace{35mm} \frac{SWIR2-Red}{(SWIR2+RED)(Green^m + Red^m + NIR^m)}
\end{equation}
\hspace{9mm} \textbf{BASMA} = \hspace{30mm} Spectral Mixture Analysis 
$
\begin{cases}
band1 &\text{green vegetation}\\
band2 &\text{soil + non-productive vegetation}\\
band3 &\textbf{char}
\end{cases}
$
\end{tcolorbox}
\caption{Burn indices used in analysis}
 \label{fig:fig2}
\vspace{-4mm}
\end{figure}
\vspace{-3mm}
\subsubsection{Separability Analysis}
To assess which sensor products and burn indices are best suited for our data and objectives, we conducted a separability analysis, using the parametric M-statistic \cite{Kaufman&Remer1994}. We compared burned plots both to unburned plots observed in the same image and to the nearest pre-burn observation, both common practices in the burned-area mapping literature \cite{Huang2016, Chuvieco2002}. To compare burned to unburned plots within the same image, we utilized our ground data to match burned rice plots to rice plots that were known to have been tilled in the same period but not burned. As the burned plots in our study are almost always tilled shortly after the burn event, burn events on such plots compared to other tilled plots are likely to be less spectrally distinct compared to other burn events that may be occurring at the same time. 
The immediate tilling event also likely obscures the burn signal more quickly than with other burn events. To assess the longevity of windows of spectral separability, we made use of the high temporal frequency of PlanetScope imagery to construct a data set of plots where the date of the burn event is known within two days, meaning that a plot had a clear not-burned observation followed by a burned observation no more than two days after. With these data, we conducted an initial pre-/post-burned matched analysis, similar to that of \cite{Huang2016}, then tested the temporal robustness of this signal by comparing with observations increasingly farther from the burn date.   
\vspace{-4mm}
\subsubsection{Feature Creation and Selection}
Recent efforts to map crop burning in India demonstrate the superior performance of models built with multiple indices and bands compared to single indices \cite{Deshpande2022}. After determining the indices with the highest potential in separating burned and unburned classes, we combined these indices and the underlying bands as features using a machine-learning approach to allow maximum exploitation of the spectral information. As the goal of the model is to detect whether a plot was burned at any point during the burn season, we stacked all overlapping images into a time-series data cube for each study plot and created features at the pixel level based on statistics from each band and derived index across time. Statistics included min, max, median, and outer percentiles (10th, 20th, 80th, 90th). An additional temporal differencing measure (Vdiff) was calculated for each band and index with the goal of capturing the moment the pixel changed from unburned to burned. This Vdiff measure was calculated based on the largest drop or spike in the sequence of values (V) for $V_{t+1}-V_{t}$. To reduce effects of noise, a buffer parameter (b) was applied such that this drop/spike was only recorded if the value remained below/above a threshold (usually the mean) for the next b images. Buffer values of zero, one and two images were applied to each Vdiff measure and the resulting variable with the highest importance score during model development was retained.

Pixels along the edge of a plot likely present differently from inner pixels due to the mixed occurrence of plot/non-plot classes within border pixels. Dropping border pixels would reduce such noise, however doing so could also reduce burn detection, as signs of burning often linger along the border after the plot has been tilled. We added a border flag to our model to assess the added benefit of using border pixels. If the border variable was given low importance in a model, the border pixels were dropped altogether from that model.

\subsection{Machine Learning Model}
We used pixel-level features from the 681 labeled plots to train a Random Forest (RF) model to provide binary burn/no burn predictions. Although the model was trained at the pixel level, full plots were held out from the training data for use in accuracy assessment. Plot-level holdouts were necessary as pixels within the same plot have highly correlated features; if some pixels within a plot were used for training while others were used for testing, overfitting of the model and overestimation of accuracy would occur. This issue could not be resolved with spatial decorrelation methods because the labels are at the level of the plot. A single plot was held out each time while a Random Forest model was generated with the remaining 681 plots. This process was repeated 681 times in a Leave-One-Out Cross-Validation (LOOCV) format. Model accuracy was assessed based on the prediction score for each plot in the run where it was left out of model training. 

We used SequentialFeatureSelector in the sklearn toolkit in Python to reduce the feature space to an optimal number of features (around 30) prior to the final LOOCV analysis. Retained features are presented in Appendix 1. The trained model was then used to provide predictions for the holdout data for accuracy assessment and threshold selection for aggregated plot-level predictions. The pixel RF to plot aggregation discussed below was then applied to the unlabeled plots to provide burn estimates for the full data set. Three separate models were evaluated: one using only Planet data, one using only Sentinel data, and one using Planet and Sentinel combined, to assess the gains from each data source. We present the combined Planet and Sentinel data as our main model; the Sentinel- and Planet-only models are included in the appendix for comparison.

\subsection{Threshold Selection for Plot-Level Aggregation}

To convert from pixel to plot-level predictions, we first aggregate from pixel-level data to plot-level data by taking the plot-level mean of the continuous RF output.\footnote{We also explored using aggregation statistics other than the mean, including the median value and percentiles ranging from the 25th to 90th. The mean reliably outperformed these other statistics.} We used two approaches to setting a classification threshold based on the mean prediction scores, with plots exceeding the threshold classified as burned. First, we maximized overall accuracy (``max accuracy'') by iterating over each threshold percentile, and selecting the threshold with the highest accuracy for the full labeled set of plots. Alternatively, to balance accuracy across burn and no-burn labels (``balanced accuracy''), we iterated the burn accuracy and the unburned accuracy over each threshold percentile, interpolated these accuracies into smooth functions, and selected the percentile threshold with the greatest accuracy for the mean at the point of intersection (where burned accuracy equals unburned accuracy). We tested using Cohen's Kappa for threshold optimization, which measures how a classifier compares when evaluated against a random classifier. In this case, maximizing kappa resulted in the same threshold selection as the max accuracy approach for all versions of our model.

Note that the RF model outputs a continuous prediction score ranging from [0, 1.0], however this measure does not represent the probability that the given pixel is burned; rather it is the proportion of decision trees that classified it as such. The default threshold that was used in our RF model for classifying between burned and unburned pixels is 0.5, where any value above 0.5 is classified as burned. This threshold is arbitrary, and represents a classification rule of a simple majority vote of the ensemble, appropriate at the pixel level. After aggregating to the plot level, this classification threshold loses some relevance. Therefore, our threshold-setting decision rules of max accuracy and balanced accuracy represent different value judgments regarding how we spread errors between burned and unburned classes. Our work follows other examples where RF models trained on imbalanced data are improved by changing the classification threshold in a post-processing step \cite{esposito_ghost_2021-1,provost_robust_2001}.


\section{Results}
\subsection{Observation Frequency and Gaps}
Plots are observed on average every 6.8 days with Sentinel-2 and 2.2 days with PlanetScope (Table \ref{windows}). Clouds and heavy haze cause missed observations, with the largest gap averaging 12 days with Sentinel imagery and eight days with PlanetScope imagery. This observational gap is exacerbated by the fact that clouds and extreme haze are more likely during the period of heaviest burning. 

\vspace{-2mm}
\begin{table}[H]
\caption{Window between observations (in days) for all study plots}
\vspace{-4mm}
\begin{center}
\label{windows}
\begin{table}[H]
\centering
\begin{tabular}{lllrrrrr}
  \hline
 &  across plots & across time & Planet & Sentinel \\ 
  \hline
 & Mean & Mean & 2.2 & 6.8 \\ 
   & Mean & Max & 8.4 & 12.2 \\ 
   & Max & Mean & 5.6 & 7.8 \\
   & Max & Max & 13.0 & 15.0 \\
   \hline
\end{tabular}
\end{table}    
\end{center}
\end{table}
\vspace{-6mm}

\FloatBarrier

\vspace{-8mm}
\begin{figure}
\begin{center}
    \includegraphics[width=17cm]{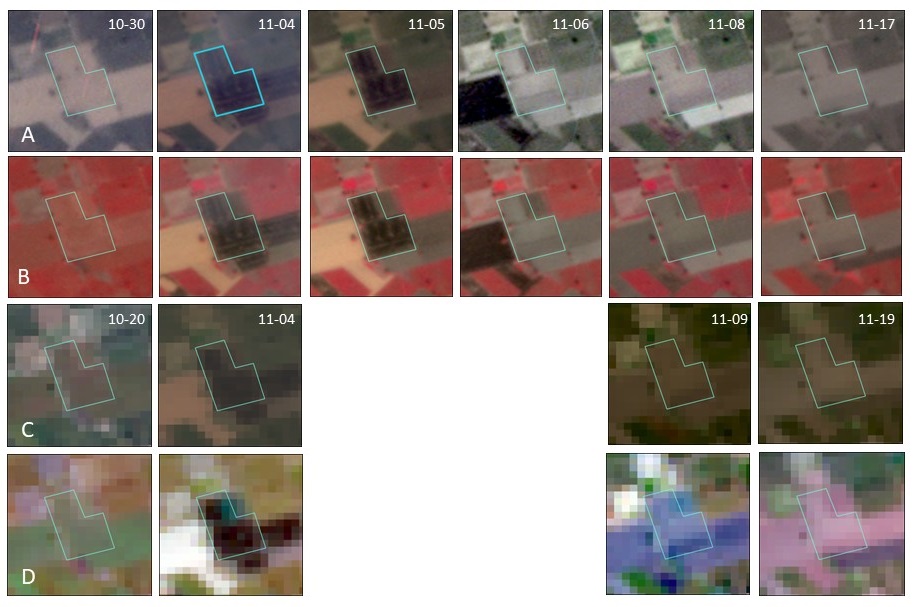}
\end{center}
\vspace{-4mm}
\caption{Example of burn observations in PlanetScope and Sentinel-2 data}
\footnotesize{Observations of a plot burned on 11-04. Row A shows PlanetScope observations viewed in true color (rgb); Row B shows the same PlanetScope observations viewed to emphasize NIR band (nir-red-blue); Row C shows Sentinel-2 observations viewed in true color (rgb); Row D shows the same Sentinel-2 observations viewed to emphasize SWIR and NIR (SWIR1-NIR-red). }
 \label{fig:fig3}
\end{figure}

\subsection{Indices and Separability}
\begin{figure}
\begin{center}
\includegraphics[width=15cm]{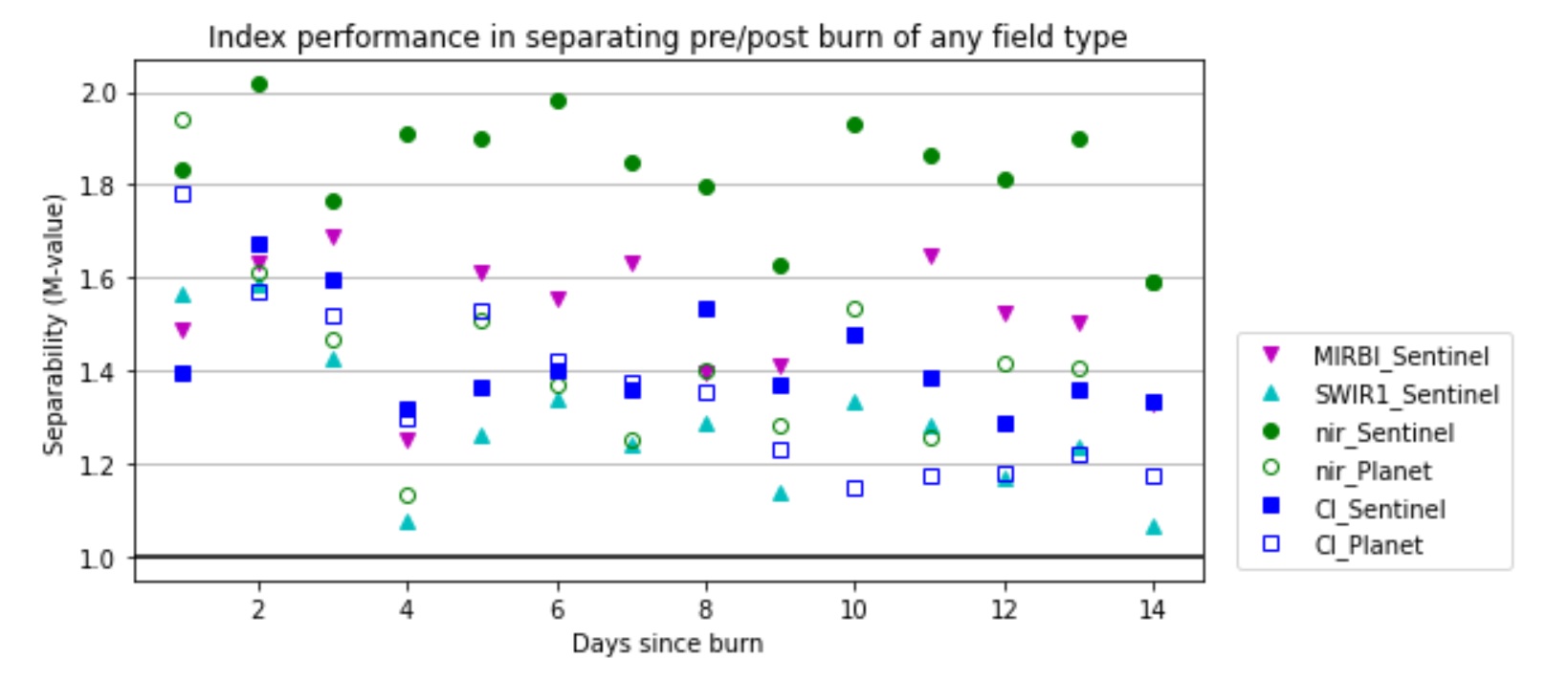} 
\end{center}
\vspace{-4mm}
\caption{Ability of indices to separate unburned/burned plots with increasing time since burn event}
\footnotesize{ M-Values > 1 indicate some separability M-values close to 2 indicate excellent separability.}
 \label{fig:fig4}
\end{figure}

If observed within two days, burn scars are quite obvious in both PlanetScope and Sentinel-2 images (Figure \ref{fig:fig3}). Separability analyses of pre- and post-burn observations, a common method of assessing index performance \cite{Perez2022, Huang2016}, indicate that several of the indices perform quite well in identifying burning (Figure \ref{fig:fig4}). The char index, based on visible bands only, performs particularly well in the initial days post-burn, but this ability to separate burned from unburned plots quickly degrades with time. BASMA performs similarly to, and possibly better than, the char-index, but could not be evaluated in the same way as the other indices due to the use of weekly mosaics, which obscures the specific date of observation. While other indices achieved an M-value above the threshold of 1.0 generally considered to indicate good separability \cite{Kaufman&Remer1994, Veraverbeke2011}, this value is somewhat arbitrary \cite{Smith2007} and top indices shown in Fig \ref{fig:fig4}, all reached an M-value greater than 1.5 at some point. The narrow detection window provided by the char index poses a particular problem in detecting burn events using satellite imagery with low temporal resolution. This ephemeral signal can be a problem for detection even with imagery with frequent revisits such as PlanetScope given that the average plot had an eight-day gap in PlanetScope observation and some plots were not observed for as long as 13 days. Separability analysis with Sentinel-2 bands shows that the extra spectral information from Sentinel-2 can increase detectability compared to PlanetScope for slightly longer intervals, but is not enough to compensate the lack of information from the event itself.

\begin{figure}
\begin{center}
\includegraphics[width=15cm]{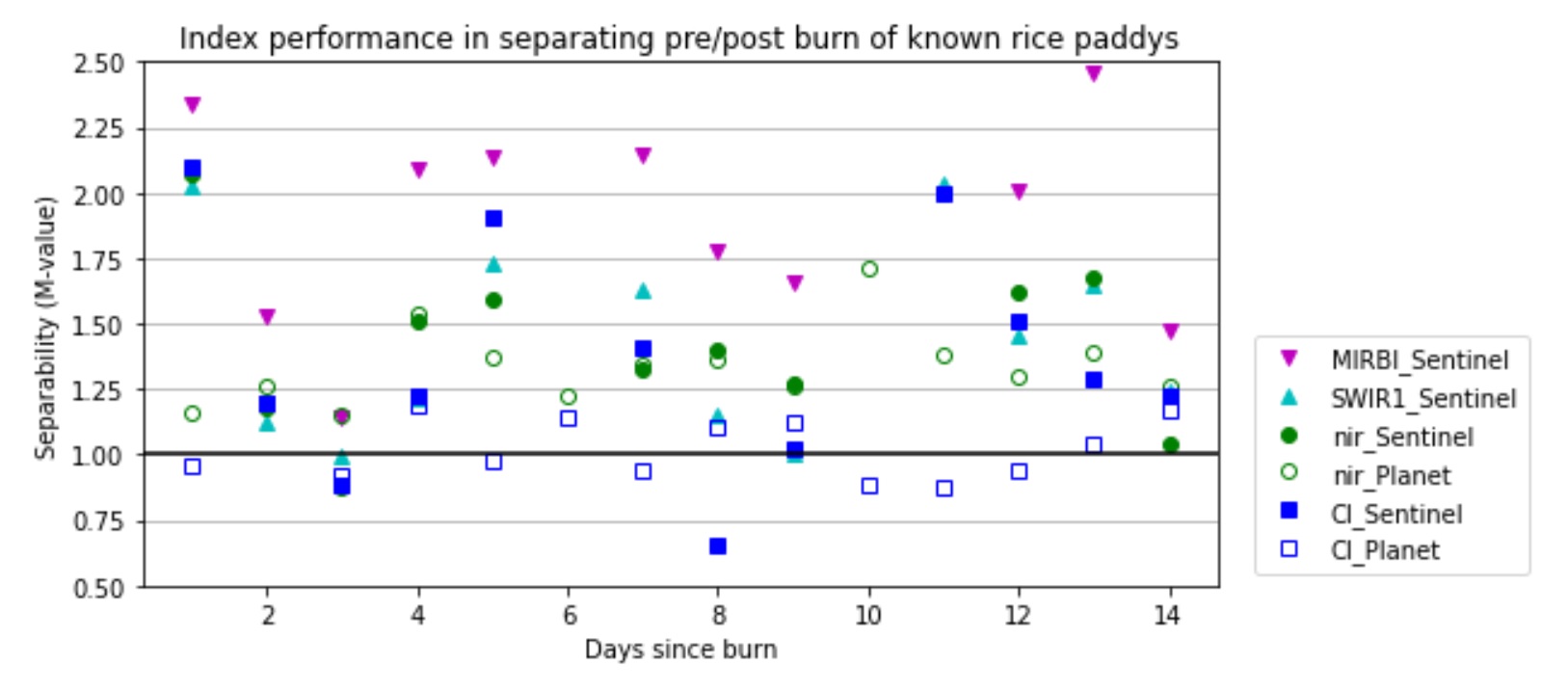}
\end{center}
\vspace{-4mm}
\caption{Ability of indices to separate unburned/burned rice paddy visited with increasing time since burn event}
\footnotesize{M-Values > 1 indicate some separability; M-values close to 2 indicate excellent separability. The higher variation seen in this figure compared to the previous is due to the small sample-size of 62 burnt plots seen at different points in time}
 \label{fig:fig5}
\end{figure}

This type of pre-/post-burn analysis provides useful information regarding the ability of indices to detect burning in optimal conditions, but it does not fully inform the task at hand, given that the burn event observed in the imagery can be any type of plot, and not necessarily rice paddy that is immediately tilled after burning like those in our study. By narrowing the sample to 62 plots from our study where a burn event is clearly observed in the PlanetScope data, we confirmed that separability is lower within a single image and dissipates even more quickly in these plots compared to our general sample, although variance is high for these 62 plots (Figure \ref{fig:fig5}).

\begin{figure}
\begin{center}
\includegraphics[width=18cm]{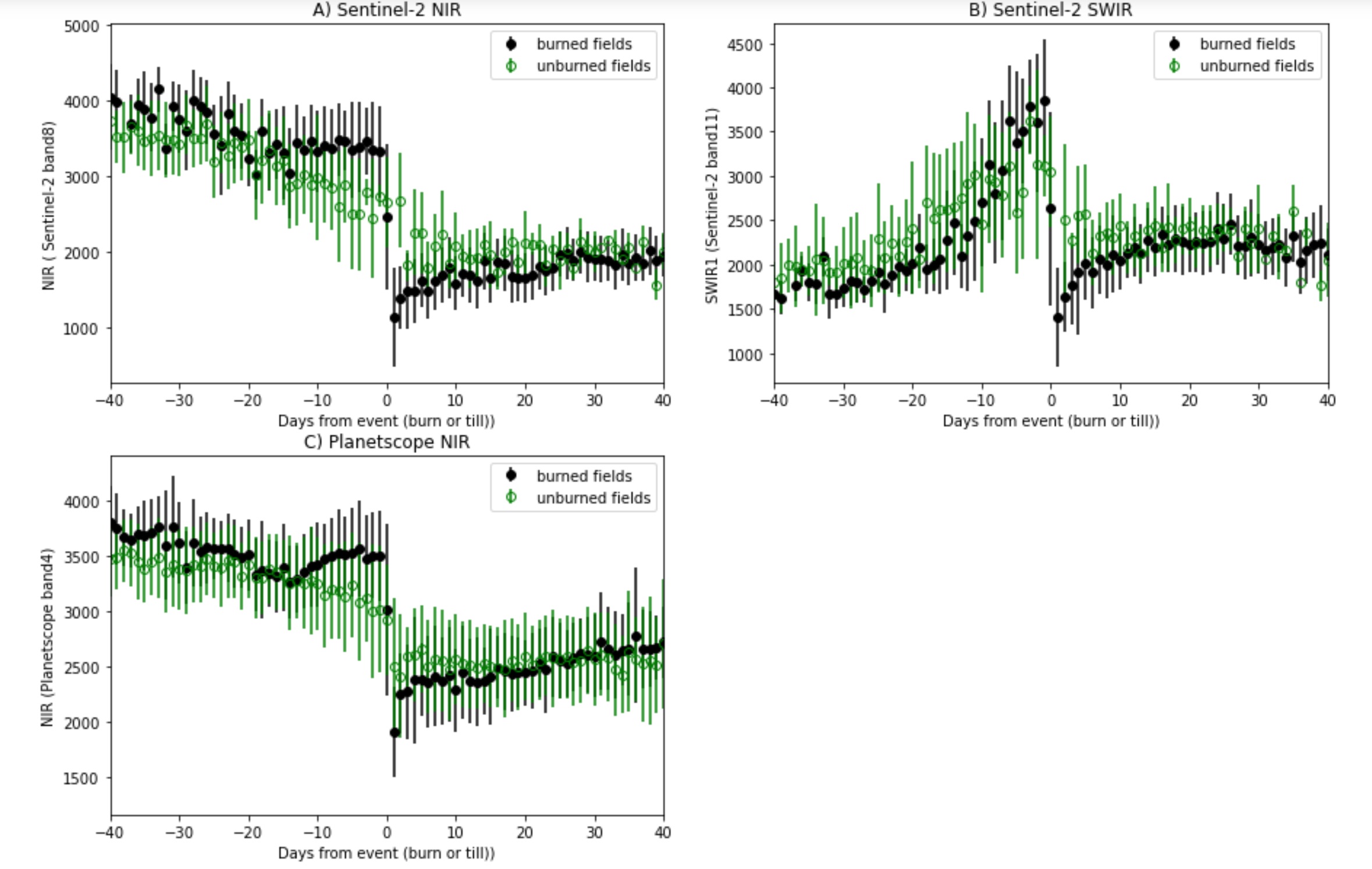}
\end{center}
\caption{Spectral signatures of burned and unburned-tilled plots across time with respect to burning/tilling event}
\footnotesize{Points and error bars are mean and standard deviation of study plots with known burn-date (n=62) or post-till visit (n=212). Burned plots have a known burn date observed in PlanetScope imagery. Unburned plots were visited on the ground soon after tilling, thus tilling has occurred by time 0 but could have occurred up to 10 days earlier.}
 \label{fig:fig6}
\end{figure}

Even after limiting the sample to the rice paddies in our study, this pre-/post-burn approach to assessing separability yields overly optimistic results. While it is clear that the separability provided by indices that detect burning based on the charring evident in the visible light bands is short lived, the near- and shortwave-infrared bands show longer-lasting separability pre- and post-burn, which can magnified by the MIRBI index. This pre-/post-burn analysis does not take into account information from the actual plots we are trying to separate from the burned plots, namely rice paddies that have been tilled but not burned. When NIR and SWIR signatures of burned plots are compared to those of the 240 plots known to be tilled but not burned, almost the same effect is observed in the unburned plots (Fig \ref{fig:fig6}). Separation of pre- and post-tilled plots can be achieved with high accuracy using the NIR and SWIR bands, but determining whether they were burned prior to tilling is less straightforward. It does appear that the NIR and SWIR bands decrease more following a burn than following a tilling event alone, but this difference does not last more than a few days. It should be noted that the exact date of tilling is not known, although it is known to have occurred prior to time zero for unburned plots in figure \ref{fig:fig5}. The lower average values for unburned plots compared to burned plots just prior to time zero are likely due to the tilling event occurring a little earlier than estimated for some plots.

 \FloatBarrier
 
\subsection{Random Forest Model and Threshold Selection}
Burns are detected with highest accuracy when information from both Sentinel and PlanetScope is combined, although the models from the individual sensors only achieve slightly lower accuracy (see Appendices A and B). The features given the most importance in all models (See Appendix C) include variables derived from indices and bands found to have the highest separability (NIR, CI, BASMA, MIRBI) as well as the basic visual bands and vegetation indices, likely providing contextual information. Only the 50 most important variables were used in the final model to reduce correlational interference. The border flag was given one of the lowest importance scores and border pixels were thus deleted from the final models. As suggested by the separability analysis in Section 3.2, frequency of observations is very important for burn detection, with the number of total Sentinel images given one of the highest importance scores; pixels observed more often by Sentinel were more likely to be classified as burned. The same is true for PlanetScope, although this variable was given lower importance overall.

\begin{figure}
\begin{center}
    \includegraphics[width=14cm]{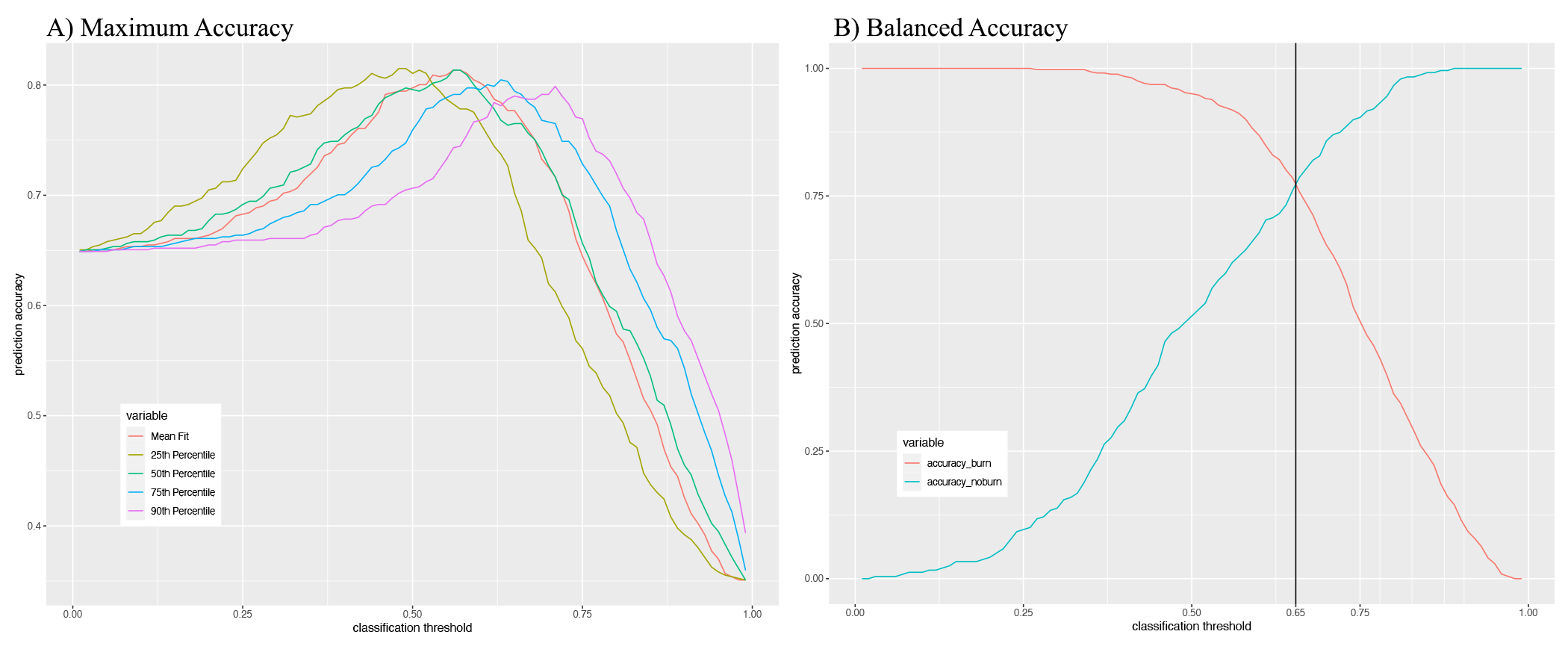}
\end{center}
\caption{Plot-level threshold selection for maximum accuracy and balanced error approaches}
\footnotesize{For maximum accuracy (A), the selected threshold is the point where total accuracy is maximized after iterating over each threshold percentile.The mean produces higher accuracies than the other percentiles tested and is thus used in all models. For balanced accuracy (B), we then determine where the burn-accuracy and no-burn accuracy curves intersect to select the final threshold}
 \label{fig:fig7}
\end{figure}

 Following plot-level aggregation, our best RF model achieves 82-percent overall accuracy, with 91-percent accuracy in detecting burned plots but only 63-percent accuracy in detecting unburned plots (Table \ref{accuracy}). When the burned/unburned errors are balanced with our balanced accuracy procedure, the overall accuracy is reduced to 78-percent.  Figure \ref{fig:fig7} shows how the aggregation threshold was selected for the max-accuracy model and balanced accuracy models.
\begin{table}[H]
\caption{Model accuracy for thresholding methods}
\begin{center}
\def\sym#1{\ifmmode^{#1}\else\(^{#1}\)\fi} \begin{tabular}{@{\hspace{2mm}}@{\extracolsep{-2pt}}{l}*{2}{>{\centering\arraybackslash}m{1.7cm}}@{}}  \toprule
 & Max accuracy & Balanced accuracy\\
\midrule
False burn & 88 & 57 \\
False no burn & 38 & 95 \\
True burn & 404 & 347 \\
True no burn & 151 & 182 \\
\midrule
No burn accuracy & 0.63 & 0.76 \\ 
Burn accuracy & 0.91 & 0.79 \\
\textbf{Mean accuracy} & \textbf{0.82} & \textbf{0.78} \\
\bottomrule
\end{tabular}    
\end{center}
 \footnotesize{Values are at the plot level and are derived from labeled plots only. False burn is the number of plots that the model predicted as burned but were labeled as not burned based on ground observations. False no burn is the number of plots that the model predicted as not burned but were labeled as burned based on ground observations.}
 \label{accuracy}
\end{table}
 
\subsection{Application of Model Predictions for Full Data Set}
When the trained models are applied to our full set of unlabeled and labeled study plots, we see that the max accuracy model predicts burning at a much higher rate than the balanced accuracy model (table \ref{summary}). Table \ref{classification} compares the model results directly and shows that the max accuracy measure obtains a higher overall accuracy by classifying 440 plots as burned (1) that are classified as not-burned (0) in the balanced accuracy. This max accuracy model is the default RF outcome, however it is not necessarily the best. The fact that burning already occurs at a higher rate than non-burning drives the model to favor this outcome and potentially overstate the amount of burning in the unlabeled data. Our balanced accuracy model takes this into account and equalizes the error for burned and unburned plots. By relabeling some burned plots as not-burned (moving the threshold to the right), this second model balances accuracy but with the trade-off of potentially understating burning and adding no-burn labels somewhat arbitrarily.

\begin{table}[H]
\caption{Summary statistics, Model output}
\vspace{-4mm}
\begin{center}
\begin{table}[H]
\centering
\begin{tabular}{rlrrrrrr}
  \toprule
 &  & fields & mean & sd & max & min \\ \midrule
 & Max Accuracy & 2879 & 0.81 & 0.39 & 1.00 & 0.00 \\ 
   & Balanced Accuracy & 2879 & 0.66& 0.47 & 1.00 & 0.00 \\ 
   & Continuous & 2879 & 0.69 & 0.16 & 0.99 & 0.03 \\ 
  \bottomrule
\end{tabular}
\end{table}
  
\end{center}
\vspace{-9mm}
\footnotesize{Notes: Summary statistics for each of the threshold selection approaches and the continuous mean of the plot-level aggregation.}
\label{summary}
\end{table}

\begin{table}[H]
\caption{Burn predictions, by model}
\begin{center}
\def\sym#1{\ifmmode^{#1}\else\(^{#1}\)\fi} \begin{tabular}{@{\hspace{2mm}}@{\extracolsep{-2pt}}{l}*{2}{>{\centering\arraybackslash}m{1.7cm}}@{}}  \toprule
 & Max: 0 & Max: 1 \\
\midrule
Balanced: 0 & 552 & 425 \\
Balanced: 1 & 0 & 1902  \\
\bottomrule
\end{tabular}
\end{center}
 \footnotesize{Notes: Max and Balanced are threshold models to convert from pixel- to plot-level predictions. 0 is not burned and 1 is burned. The cells report the number of plots in each category. }
 \label{classification}
\end{table}

\begin{figure}
    \begin{center}
    \includegraphics[width=10cm]{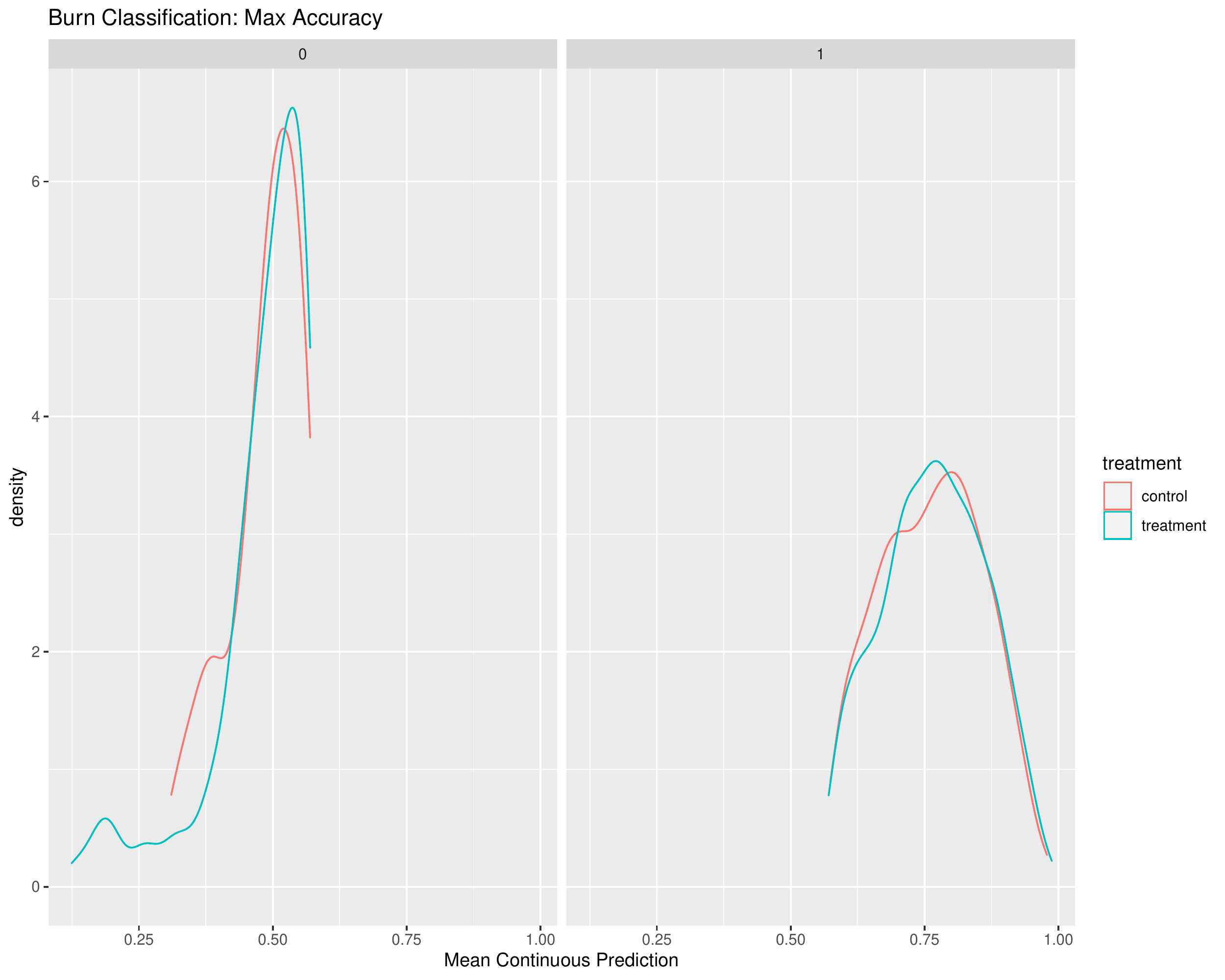} \includegraphics[width=10cm]{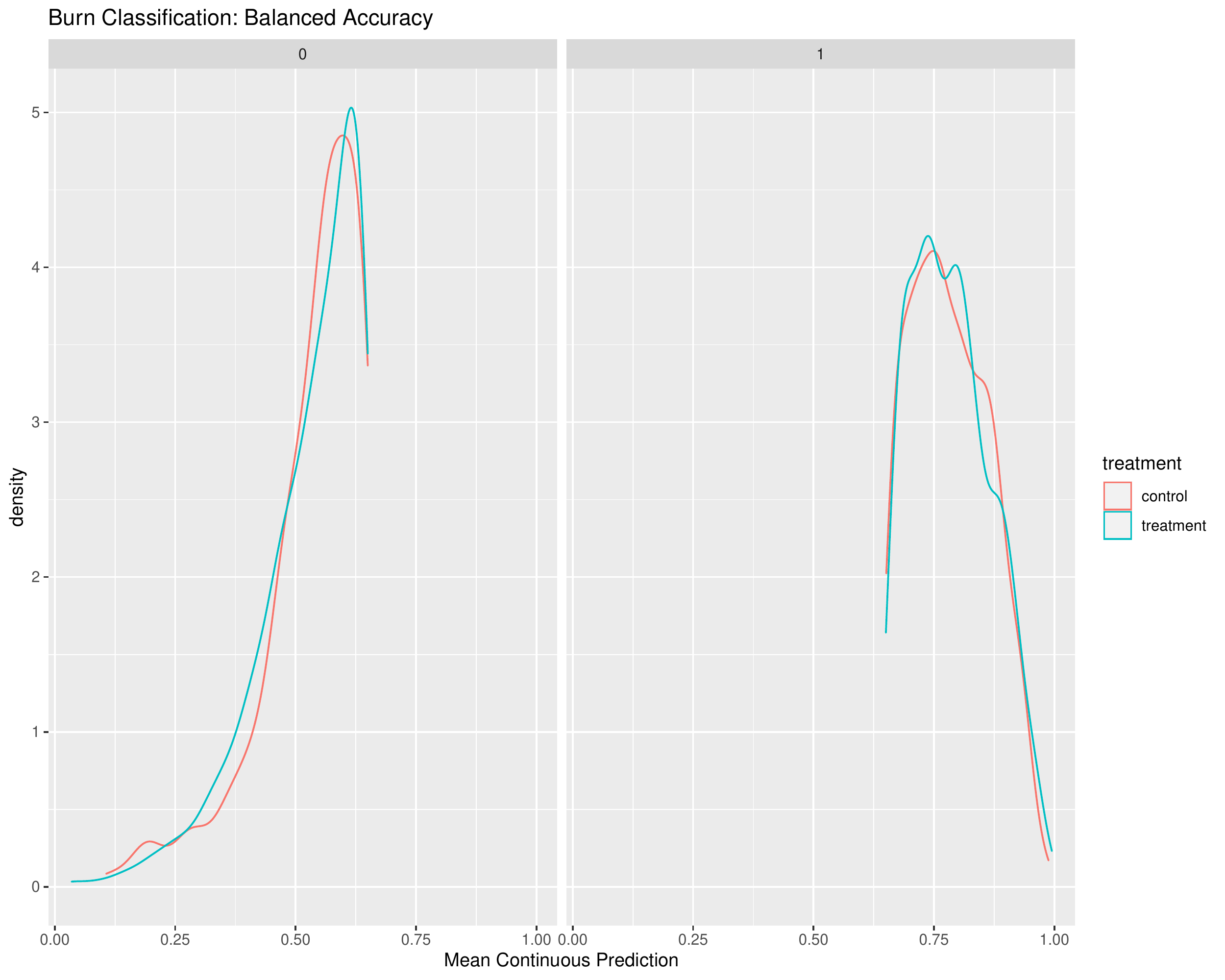}
    \end{center}
\caption{Density plots of predictions for unlabeled plots by treatment status, burn classification and threshold model}    
\footnotesize{Top panel displays the max accuracy approach; Bottom panel displays the balanced errors approach; Left side displays not-burned classification (0); Right side displays burned classification (1); Red corresponds to control plots for RCT; Blue corresponds to treatment plots for RCT. }
 \label{fig:fig8}
\end{figure}

We examined predictions for only unlabeled plots to investigate whether the distribution of model output differs by classification for treatment and control plots, based on the intervention in the broader randomized control trial (RCT). The restriction to non-labeled plots was necessary because, due to the ground-verification process, not-burned labels are available only for plots in the treatment group. In examining the density plots for the two classifications by treatment group (Fig \ref{fig:fig8}), plots classified as burned look similar in treatment and control. However, in the not-burned plots, slight differences are observed. Especially in the max accuracy model, there is a higher density of control plots than treatment plots that are classified as not-burned with a value close the threshold, indicating ambivalence between burned/not-burned. Similarly, non-burned has a longer left tail for treatment plots with both the max accuracy and balanced accuracy models, indicating that there are more plots in the treatment set that are classified as not-burned with high confidence. These observations are subtle and also opposite of the selection we are most concerned about, which is a model-driven bias toward no-burn for treatment plots. It should be noted that many of the treatment group's actual not-burned plots are likely to be in the labeled dataset (so excluded from figure \ref{fig:fig8}).\footnote{An additional four plots are excluded from figure \ref{fig:fig8} because they were not in the final RCT sample and are missing the treatment assignment used to construct the figure.}

\section{Discussion}
Distinguishing burned from unburned but tilled rice paddy plots presents many complications that are not as prominent in other burn mapping tasks. While simple vegetation indices such as SR and NDVI often perform well in burn mapping tasks \cite{Amos2019}, they are not relevant to detection of crop-residue burning due to similar vegetation changes occurring in both burned and unburned tilled plots. Similarly, changes in near- and short-wave infrared bands, while marked between pre- and post-burned plots, are similar between pre- and post-tilled plots. Our ground data confirming plots that have been tilled without burning allows us to assess separability between these classes in a way that most crop-burn analyses cannot. 

While differences can be detected between burned and unburned plots using the short-wave infrared band and char indices based on visible bands or a combination of visible and infrared bands, theses differences in signals are very short lived. A plot must be observed within days of a burn event for accurate detection of burning. Given this and the frequent gaps even in PlanetScope data, our overall model accuracy of 82-percent is respectable. The RF model with this level of accuracy selected features from the NIR band, char-index, and MIRBI among the most important variables, along with simple red, green, and blue band data that likely informs baseline conditions \cite{Deshpande2022}.

While 82-percent model accuracy might be in the ballpark of what is achievable for this task, our model does present some imbalances that, while common to this type of mapping, should be flagged for caution, especially in the context of the objective of impact evaluation. The dilemma of imbalanced errors, which occurs when one outcome (burning, in this case) occurs more frequently than the other (not burning), is common in such tasks \cite{Garbely2022,Haixiang2017} and was a topic of careful consideration in this work.\footnote{In our case, burning is observed more frequently than not burning; in most fire mapping cases, burning is usually considered the rare event.} Our approach does not solve the issue but rather provides bounds within which the solution likely lies. Moving the classification threshold provides a choice regarding how we should balance the sensitivity, specificity, and total accuracy of our classification output. Other options to address imbalanced collection of burn and no-burn labels include methods such as Balanced Random Forest, which uses a cost function or weighting of classes to compensate for the underrepresented unburned class \cite{chen2004}. In our case, the elaborate verification scheme for the randomized control led trial resulted in enough verified unburned observations that class balance is satisfactory, albeit not ideal. 

A related concern with our model design, which has received even less consideration in the general literature, is the potential for bias in our classifier arising from the relative ease in collecting validation data for burned plots compared to unburned plots. Observations of burning allow for a plot to be confirmed as burnt, but a lack of observed burning at a specific point in time does not allow for a plot to be confirmed as never burned. In our case, we did collect very careful no-burn labels for 240 plots, but the systematic way in which this was collected, compared to the more random collection of burn labels, resulted in no-burn labels only being collected for the treatment arm and not the control.

To overcome the difficulties of collecting balanced and unbiased labels, one might be tempted to turn to unsupervised learning techniques. In the case of burn mapping, however, such techniques suffer at least as much as ground validation from the inability to provide reliable observation of unburned plots. If a plot is observed by the unsupervised learner to be burnt in any image, it can be classified as burnt, but if it is not observed to be burned in any image, this does not mean that it was never burned, only that such an event was not observed. Our analysis of the longevity of separability provided by different indices following a burn event, and the fact that less than 100 of the 441 plots verified fieldwork as burned were observed in the PlanetScope data within one day of burning, underscores the likelihood that such an approach underestimates burning. This also implicates the common procedure of using higher resolution imagery to validate burn mapping; unless the temporal resolution is near-daily, smallholder crop burning is likely underestimated and model accuracy overestimated with such a method.  

Beyond these challenges, our model has room for improvement, particularly in the areas of data structure and burn index development. Optimizing model parameters and variable selection is inhibited in our model due to the use of features and predictions at the pixel level, while labels were only available at the plot level. This model format was chosen based on favorable results in accuracy compared to pre-aggregated plot-level models we tested. While the pixel-level model provides better results even when not optimally tuned, further tuning of the model is difficult given that all the pixels within a plot share highly correlated characteristics and a single plot-level label. Variable importance is biased toward the most correlated variables within each plot; if the RF model can find a correlation with other pixels in a plot and a burn label is already assigned to one of the pixels, it can classify based off of that correlation. This can lead the RF model to favor variables that are not clearly correlated with burn outcome. For example, plot size, when included in an early version of the RF model, drove most of the predictions. This is also likely the reason that a border flag variable is given very low feature importance and naturally dropped from the model. If pixels were modeled independently, border position would be expected to have more influence on the model. Future work could include developing a RF model with independent pixels while retaining the information of all pixels within each plot and the ability to aggregate this information to the plot level without diluting signals. 

As an alternative to a pixel-level model, there is potential to develop a plot-level model that performs better than the original test models. A recurrent neural network (RNN) or convolutional neural network (CNN) could be built at the plot-level to exploit the ground information given the known windows of burning/tilling events. With our ground verification, there is concern that such a model would perform worse in out-of-sample prediction given that plot inspections did not occur randomly across time and our verification of non-burned labels (plots that are never burned) are concentrated within a two-week period. These limitations could be overcome with different ground measurement protocols.
 
Other potential directions for extension or improvement include developing a burn index that is more specific to our task of mapping crop residue burning and detecting burning+tilling vs just tilling. Given the importance of detecting crop burning throughout the world, it is remarkable that indices designed for wildfire detection and calibrated in very specific environments, such as the Spanish Mediterranean, continue to dominate the literature on crop-burn detection. The less common char index was found to perform much better than more common burn indices for this task, but this could likely be better calibrated for our study area and question.

Our attempt to develop a more calibrated index using the Burned Area Spectral Mixture Analysis (BASMA) approach \cite{Daldegan2019}, produced promising results but needs further tuning. Like the char index, BASMA performs very well when a plot is observed in the satellite imagery within a day or two of a burn event, but poorly after a few days of the burn event. As BASMA was specifically designed for Sentinel imagery, the lower temporal resolution results in many missed events. Furthermore, BASMA seems to be more sensitive to haze than the char index, and thus has more missed observations due to haze, which is particularly problematic during the period of most intense burning. These issues could be addressed and the power of BASMA leveraged by applying the same process to the to PlanetScope imagery to take advantage of the higher temporal resolution. Additionally, to leverage the spectral information of Sentinel, we could modify the Principal Component Analysis (PCA) method used to select the representative weekly endmembers in BASMA. This method currently relies heavily on the concept of an ideal ``burn/charred soil'' spectral signal, but could instead be tuned to the ``post-burn'' signal of a plot up to five days since burn. Through both modification of the endmember target and application to imagery with higher temporal resolution, BASMA could be a powerful tool in detecting burn events.

Despite ample room for improvement, our model and model-building process make significant contributions to efforts to estimate crop-residue burning from satellite imagery. Our ground-validated burn labels enable us to assess the frequency at which burn events are missed even by satellite imagery with very high temporal and spatial resolution, while our ground-validated missed-burn labels, a rare asset for this type of mapping, allow us to confidently assess the accuracy of our model. As we continue this research agenda, we hope to not only develop methods by which to more accurately detect crop burning, but also to open up a discussion on the need for more systematic collection of ground-validated data to expand the collective effort to measure crop-residue burning and the impact of policies to curb such burning.

\bibliographystyle{unsrt}  
\bibliography{PunjabBurn_for_Arxiv}  
\newpage

\appendix
\FloatBarrier
\section{Results for 2019 Model: Sentinel Only}

\begin{table}[H]
\caption{Sentinel-only Accuracy assessment}
\begin{center}
\def\sym#1{\ifmmode^{#1}\else\(^{#1}\)\fi} \begin{tabular}{@{\hspace{2mm}}@{\extracolsep{-2pt}}{l}*{3}{>{\centering\arraybackslash}m{1.7cm}}@{}}  \toprule
 & Pixel, max accuracy & Pixel, balanced accuracy\\
\midrule
Mean accuracy & 0.81 & 0.75 \\
False burn & 102 & 54 & \\
False no burn & 28 & 118 \\
True burn & 454 & 364 \\
True no burn & 104 & 152 \\
No burn accuracy & 0.51 & 0.74 \\ 
Burn accuracy & 0.94 & 0.76 \\
\bottomrule
\end{tabular}    
\end{center}
 \footnotesize{Notes: Confusion matrices and accuracy statistics for each of the two measures. The confusion numbers are plot-level and come from the testing data only. Note that for the pixel level models, the hold out sample is constant. False outcomes occur when the model differs from the label, e.g., False burn means that the model predicted the plot was burned but it is labeled not burned.  }
\end{table}

\begin{table}[H]
\caption{Sentinel-only model plot assignments, by measure}
\begin{center}
\def\sym#1{\ifmmode^{#1}\else\(^{#1}\)\fi} \begin{tabular}{@{\hspace{2mm}}@{\extracolsep{-2pt}}{l}*{2}{>{\centering\arraybackslash}m{1.7cm}}@{}}  \toprule
 & Max: 0 & Max: 1 \\
\midrule
Balanced: 0 & 304 & 604 \\
Balanced: 1 & 0 & 1934  \\
\bottomrule
\end{tabular}    
\end{center}
 \footnotesize{Notes: Max and Balanced are pixel level burning measures (0 is not burned and 1 is burned). The cells report the number of plots in each category. }
\end{table}

\begin{figure}[H]
\caption{Sentinel-only threshold selection for max accuracy approach}
\begin{center}
        \includegraphics[width=10cm]{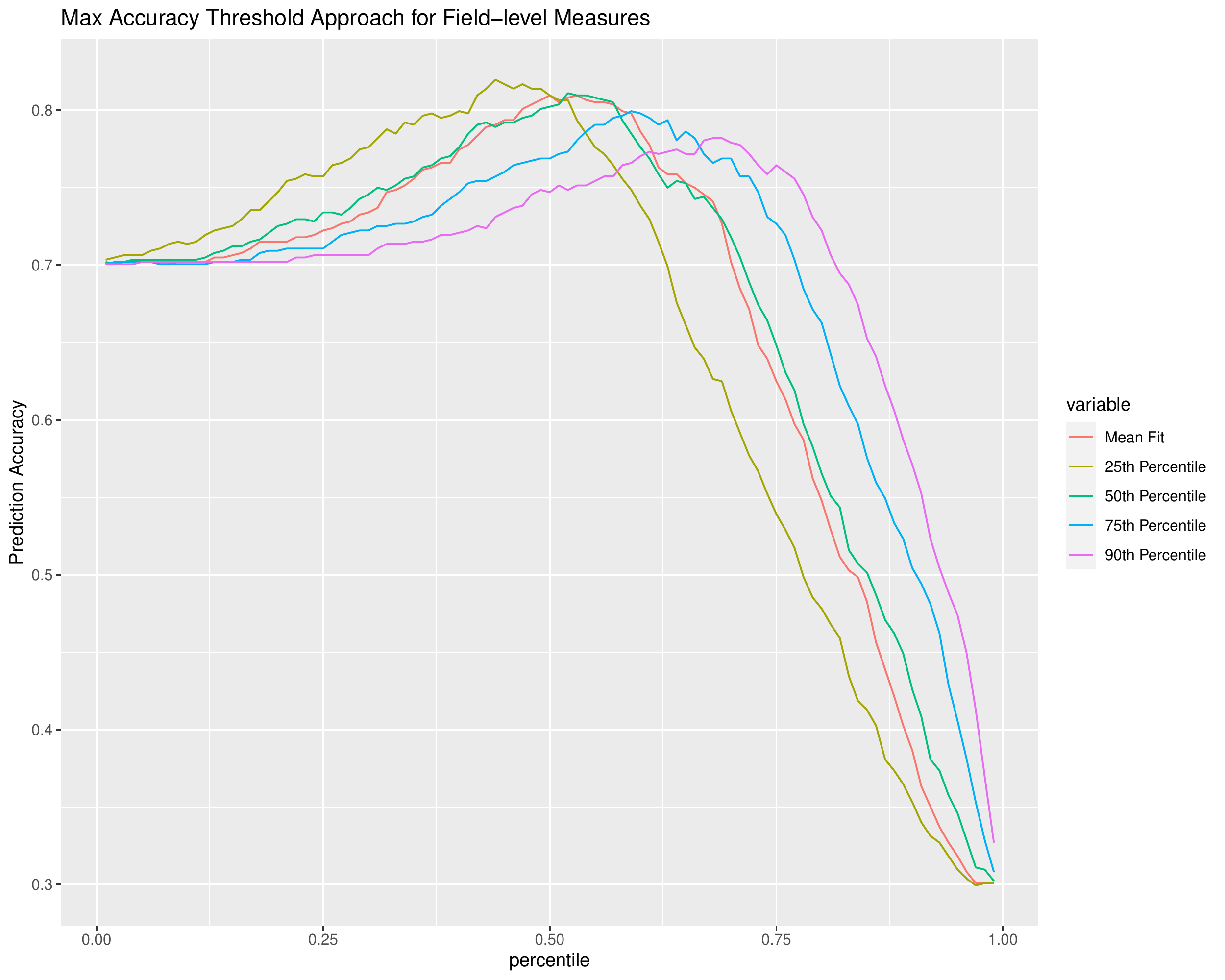}
\end{center}
\footnotesize{Notes: Pixel-level model output is summarized to the plot-level by the mean, 25th, 50th, 75th, and 90th percentile. Total accuracy is iterated over each threshold percentile. We select the threshold at which total accuracy is maximized for the mean measure.}
\end{figure}

\begin{figure}
\caption{Sentinel-only threshold selection for balanced errors approach}
\begin{center}
    \includegraphics[width=10cm]{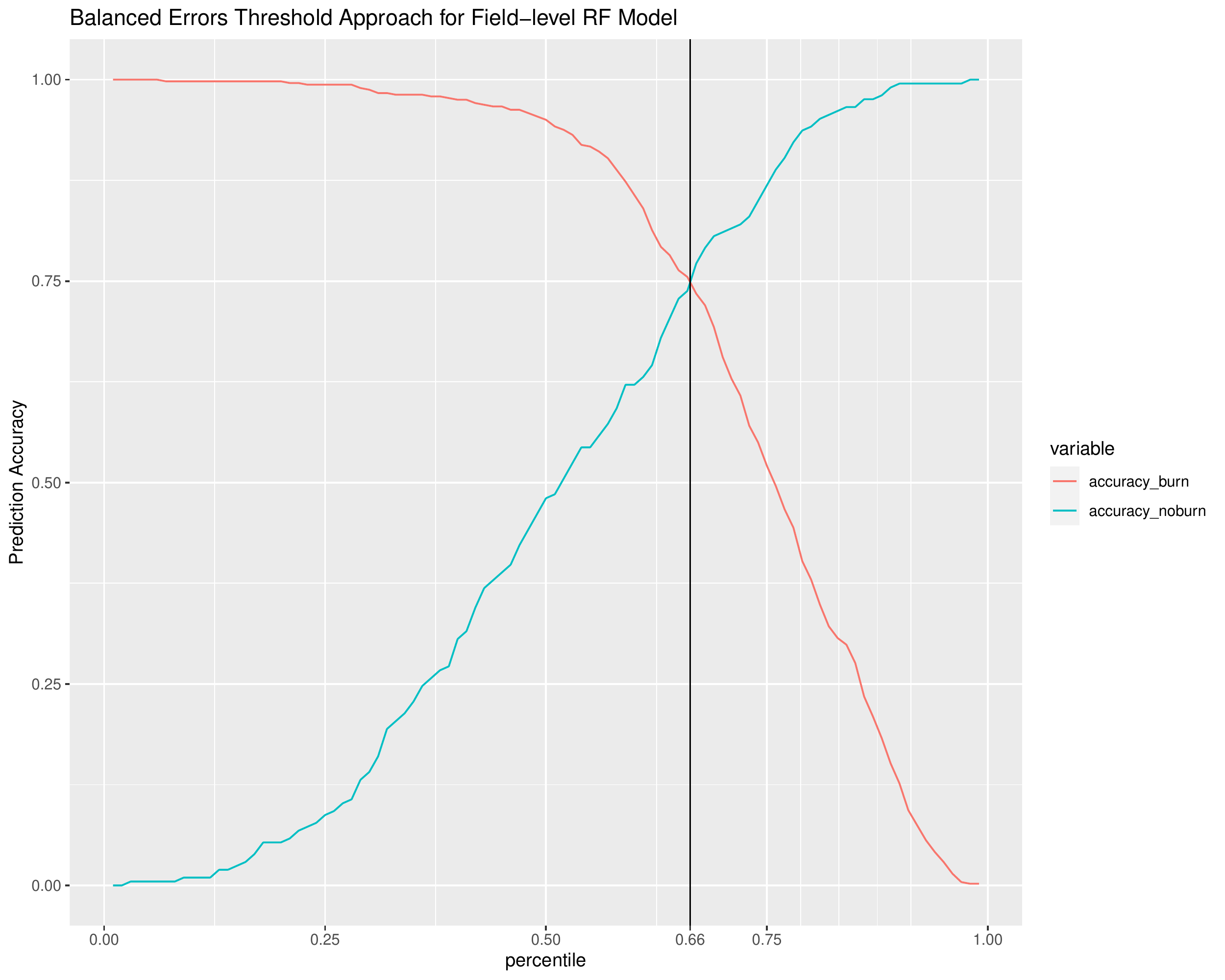}
\end{center}
\footnotesize{Notes: Pixel-level model output is summarized to the plot level by the mean. The burn and no-burn accuracy are iterated over each threshold percentile. We select the threshold where these curves intersect.}
\end{figure}

\begin{figure}
\caption{Density plots of continuous model output split across burn classification for max accuracy and balanced errors approaches}
    \begin{center}
    \includegraphics[width=4in]{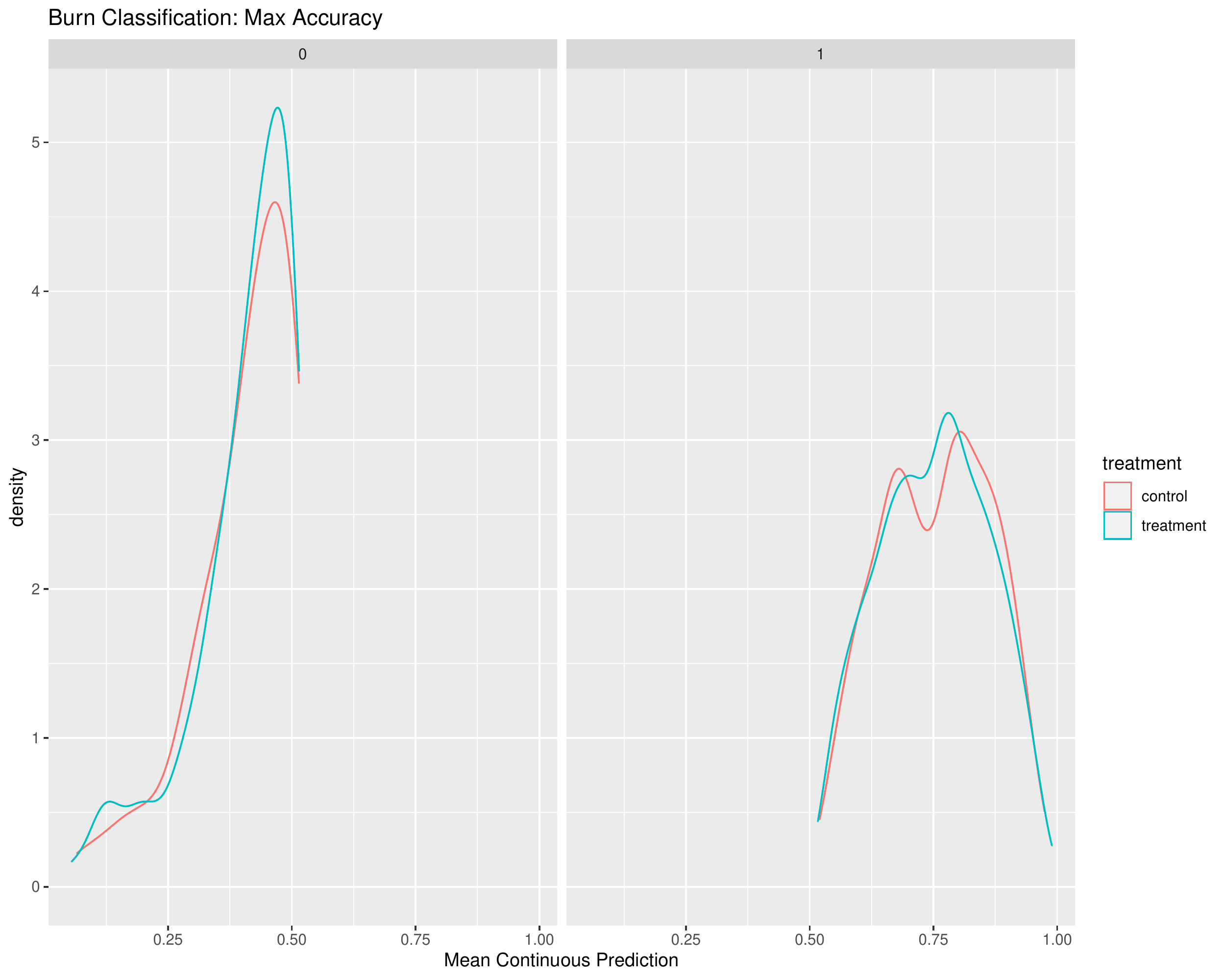} \includegraphics[width=4in]{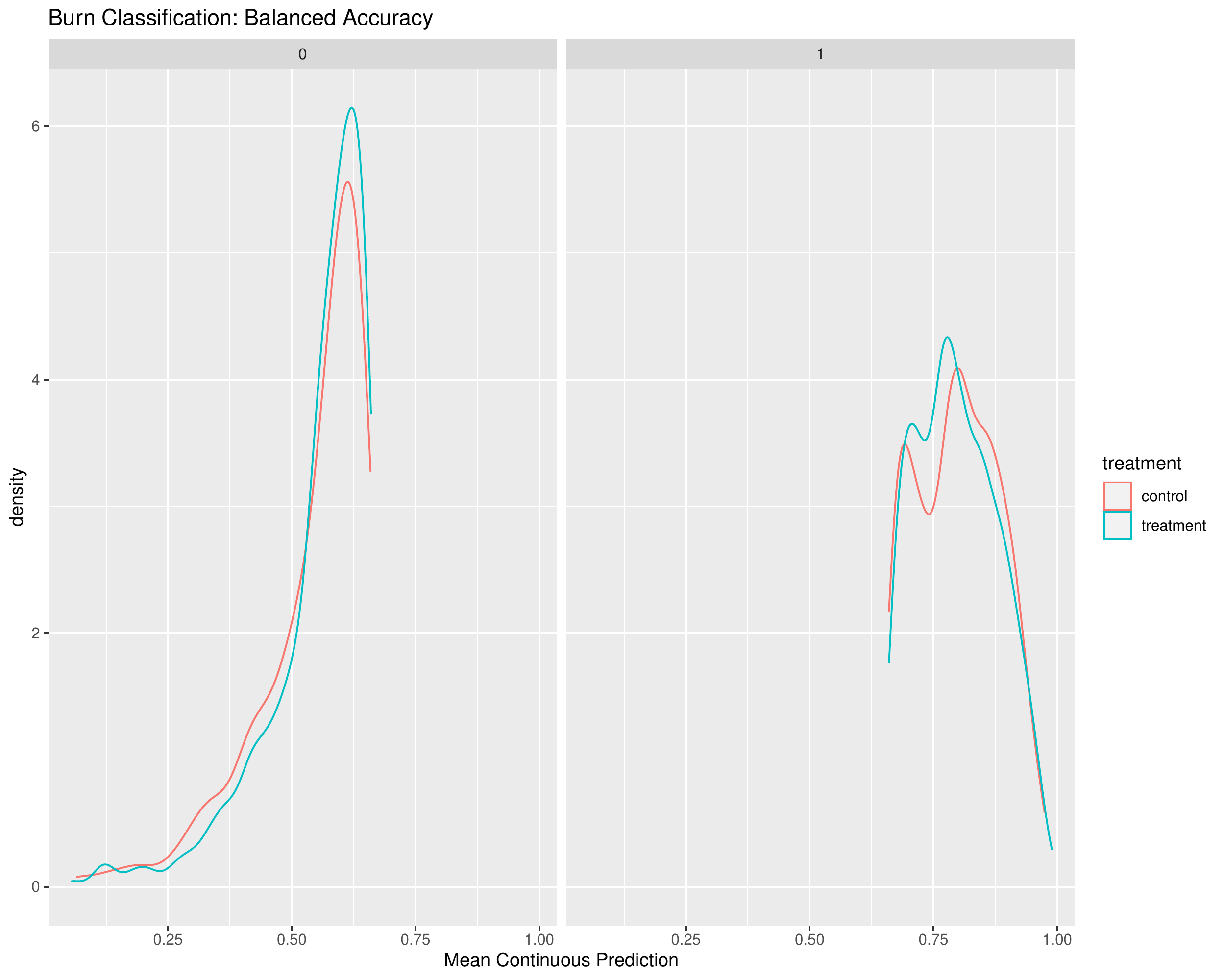}
    \end{center}
\footnotesize{Notes: Figures plot the distribution of plot level aggregates of continuous model output (pixel level), for unlabeled plots, by classification. The top figure uses the max accuracy approach and the bottom figure uses the balanced errors approach, separated by plots classified as not burned on the left and burned on the right.}
\end{figure}
\FloatBarrier
\newpage

\FloatBarrier
\section{Results for 2019 Model: Planet Only}

\begin{table}[H]
\caption{Planet-only Accuracy assessment}
\begin{center}
\def\sym#1{\ifmmode^{#1}\else\(^{#1}\)\fi} \begin{tabular}{@{\hspace{2mm}}@{\extracolsep{-2pt}}{l}*{3}{>{\centering\arraybackslash}m{1.7cm}}@{}}  \toprule
 & Pixel, max accuracy & Pixel, balanced accuracy\\
\midrule
Mean accuracy & 0.78 & 0.72 \\
False burn & 103 & 57 & \\
False no burn & 47 & 134 \\
True burn & 435 & 348 \\
True no burn & 103 & 149 \\
No burn accuracy & 0.50 & 0.72 \\ 
Burn accuracy & 0.90 & 0.72 \\
\bottomrule
\end{tabular}    
\end{center}
 \footnotesize{Notes: Confusion matrices and accuracy statistics for each of the two measures. The confusion numbers are plot-level and come from the testing data only. Note that for the pixel level models, the hold out sample is constant. False outcomes occur when the model differs from the label, e.g., False burn means that the model predicted the plot was burned but it is labeled not burned.  }
\end{table}

\begin{table}[H]
\caption{Planet-only model plot assignments, by measure}
\begin{center}
\def\sym#1{\ifmmode^{#1}\else\(^{#1}\)\fi} \begin{tabular}{@{\hspace{2mm}}@{\extracolsep{-2pt}}{l}*{2}{>{\centering\arraybackslash}m{1.7cm}}@{}}  \toprule
 & Max: 0 & Max: 1 \\
\midrule
Balanced: 0 & 449 & 593 \\
Balanced: 1 & 0 & 1798  \\
\bottomrule
\end{tabular}    
\end{center}
 \footnotesize{Notes: Max and Balanced are pixel level burning measures (0 is not burned and 1 is burned). The cells report the number of plots in each category. }
\end{table}

\begin{figure}[H]
\caption{Planet-only threshold selection for max accuracy approach}
\begin{center}
        \includegraphics[width=10cm]{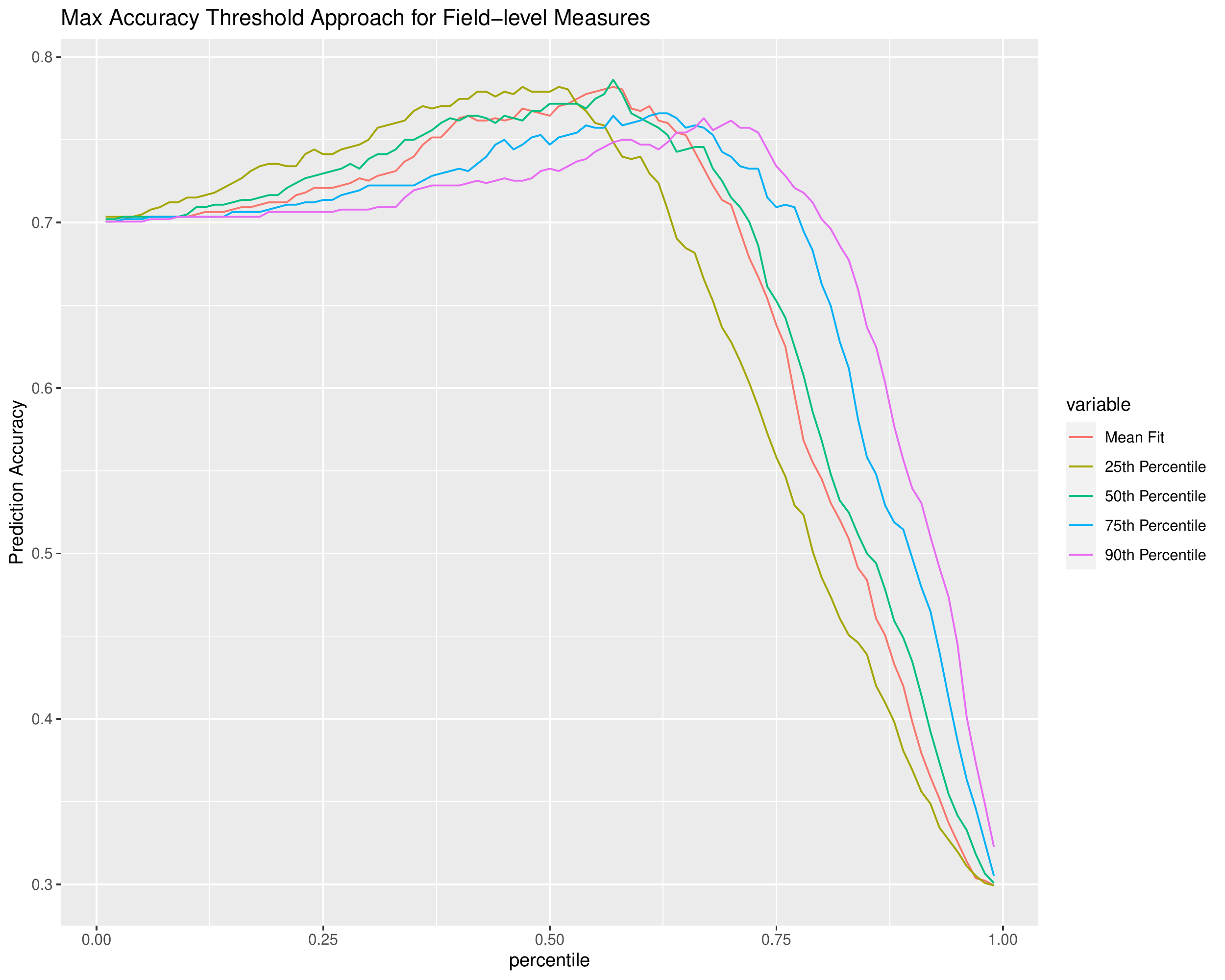}
\end{center}
\footnotesize{Notes: Pixel-level model output is summarized to the plot-level by the mean, 25th, 50th, 75th, and 90th percentile. Total accuracy is iterated over each threshold percentile. We select the threshold at which total accuracy is maximized for the mean measure.}
\end{figure}

\begin{figure}
\caption{Planet-only threshold selection for balanced errors approach}
\begin{center}
    \includegraphics[width=10cm]{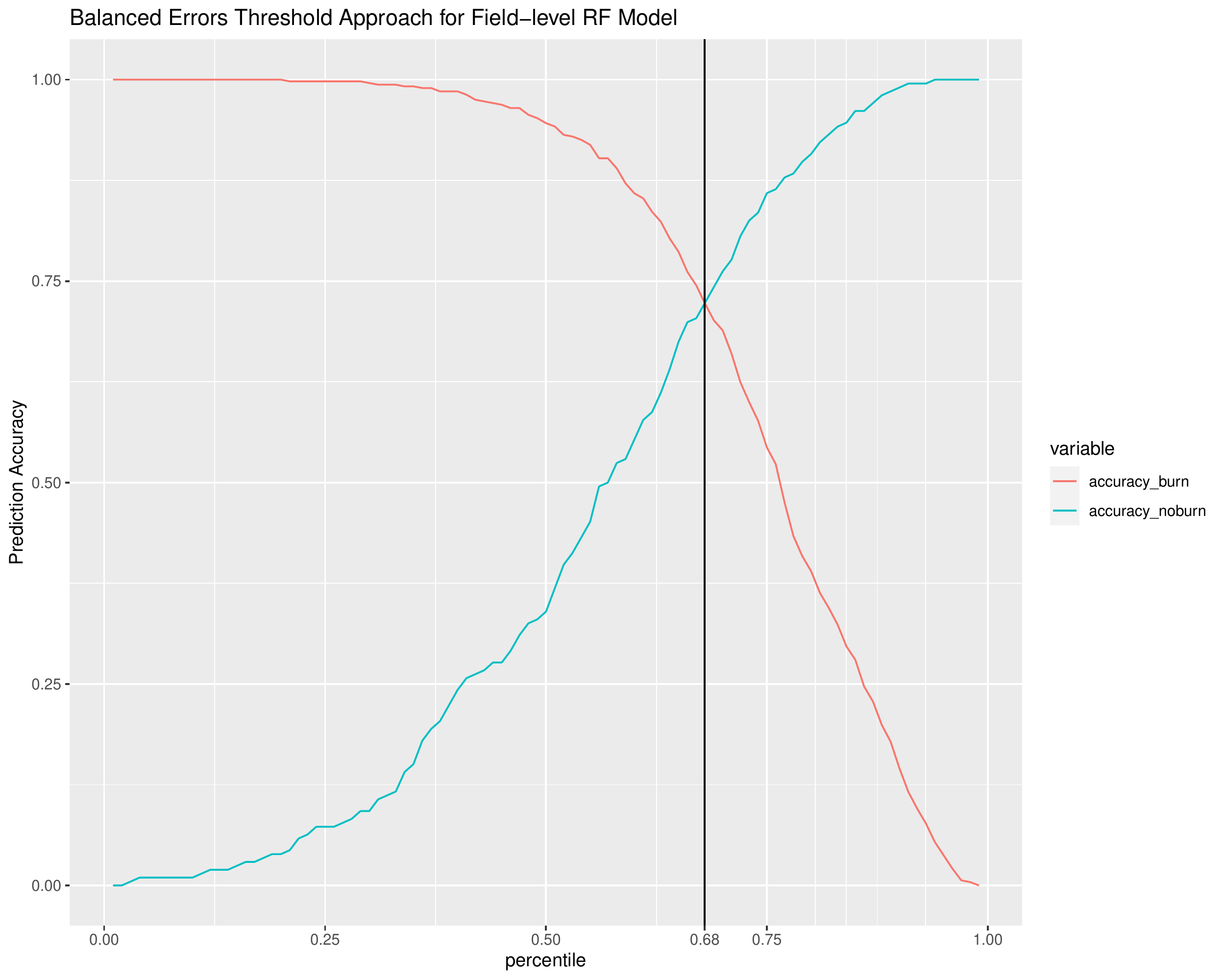}
\end{center}
\footnotesize{Notes: Pixel-level model output is summarized to the plot level by the mean. The burn and no-burn accuracy are iterated over each threshold percentile. We select the threshold where these curves intersect.}
\end{figure}

\begin{figure}
\caption{Density plots of continuous model output split across burn classification for max accuracy and balanced errors approaches}
    \begin{center}
    \includegraphics[width=4in]{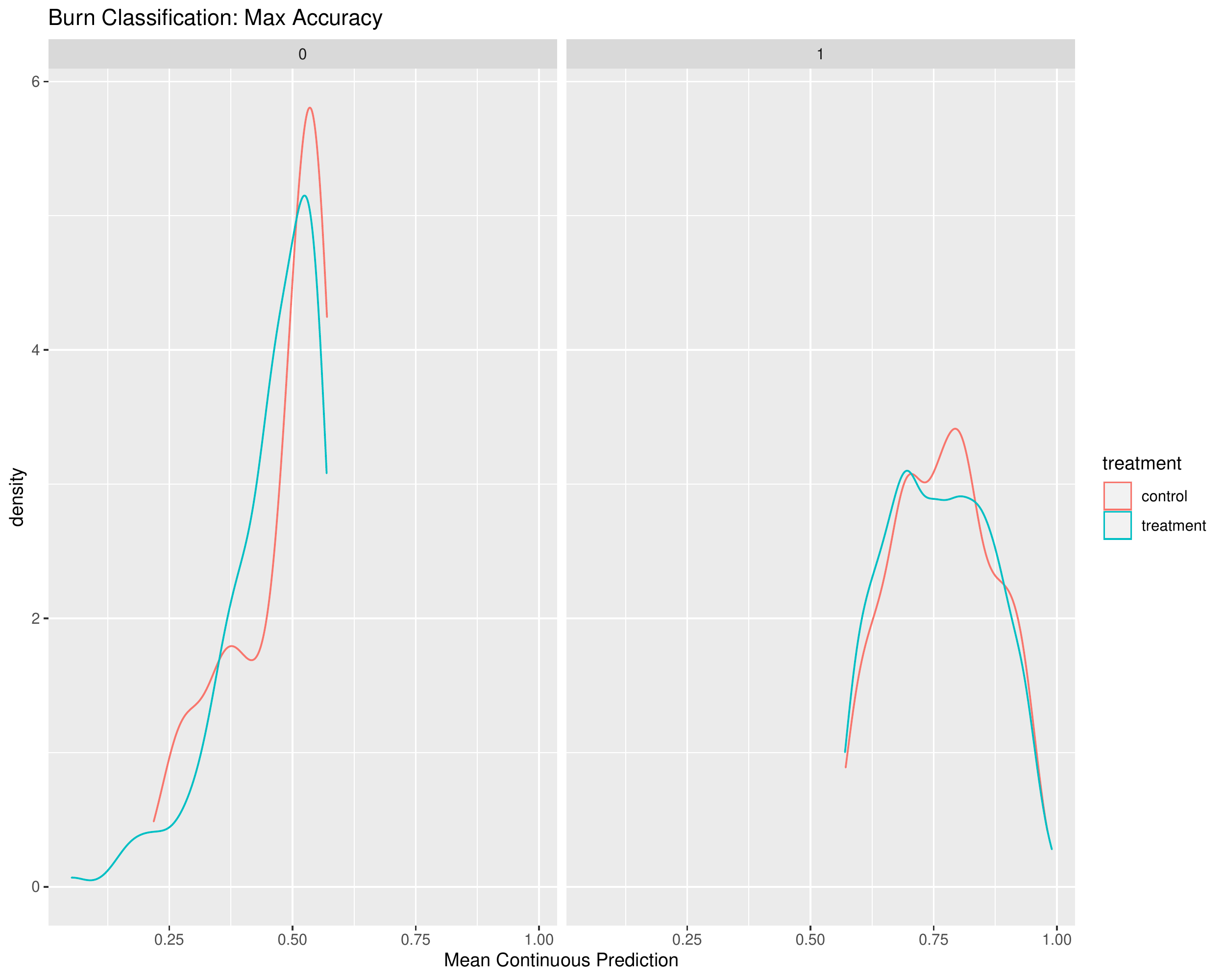} \includegraphics[width=4in]{figures/extra/V4_2019_PlanetOnly_thresholds_balAcc.pdf}
    \end{center}
\footnotesize{Notes: Figures plot the distribution of plot level aggregates of continuous model output (pixel level), for unlabeled plots, by classification. The top figure uses the max accuracy approach and the bottom figure uses the balanced errors approach, separated by plots classified as not burned on the left and burned on the right.}
\end{figure}
\FloatBarrier

\newpage
\section{Variable importance for RF models}
\begin{table}[H]
\caption{Variables retained in combined pixel-level model based on their relative importance in informing the model}
\includegraphics{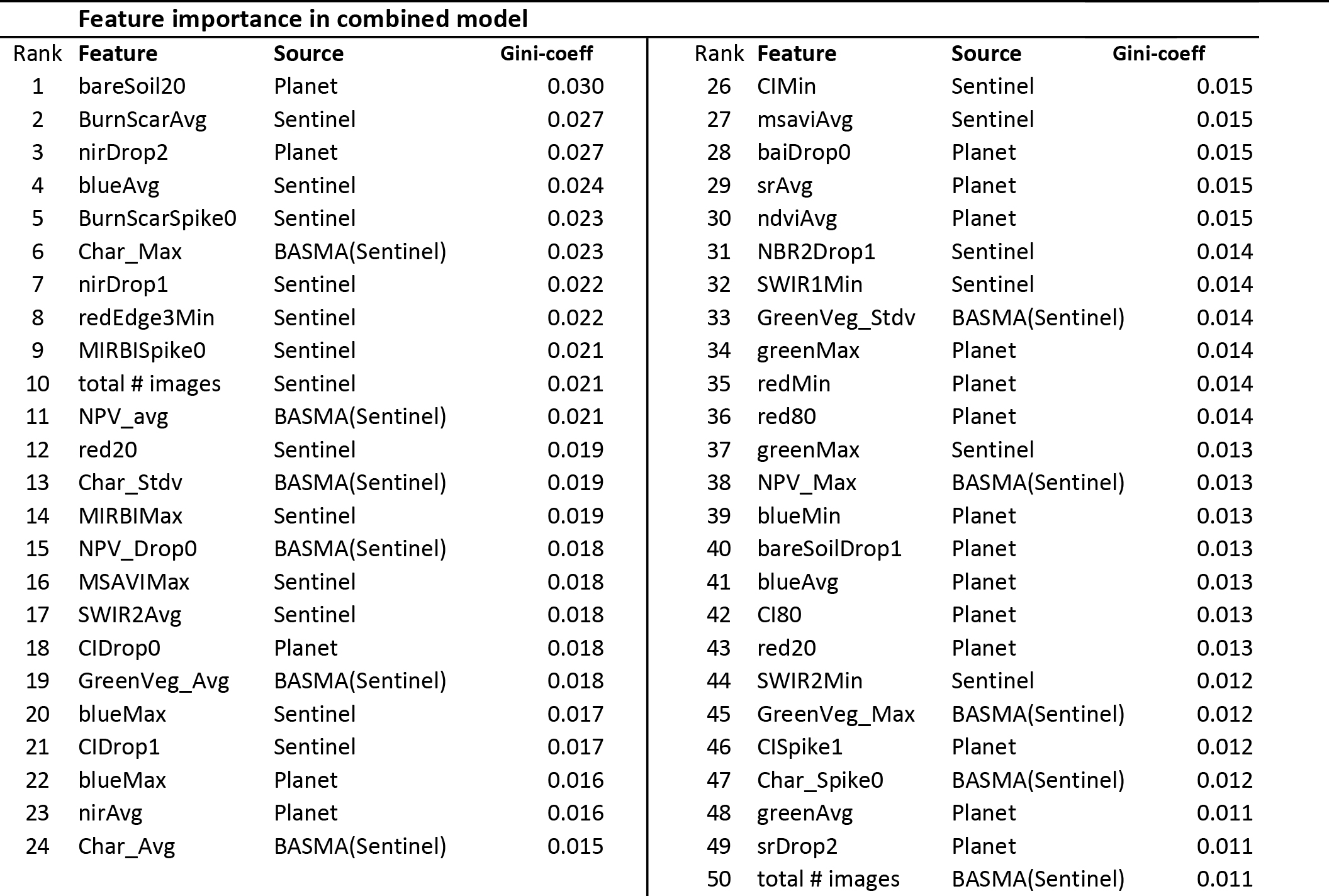}
\footnotesize{Notes: Drop0/Drop1/Drop2 is the maximum difference (Vt+1 - Vt) in the negative direction. Spike0,Spike1/Spike2 is the maximum difference (Vt+1 - Vt) in the positive direction. 0/1/2 indicates the number of subsequent images that must be above the mean for a drop/spike to be registered. MSAVI and NDVI are vegetation indices. All other indices are listed in Fig 2}
\end{table}

\begin{table}
\vspace{-40mm}
\caption{Variables retained in separate Sentinel and Planet pixel-level models based on their relative importance in informing the model}
\begin{center}
\def\sym#1{\ifmmode^{#1}\else\(^{#1}\)\fi}
\begin{tabular}{l c c l c} \toprule
 \textbf{Sentinel variables} & &    & \textbf{Planet variables}\\
 & Gini-importance &    & & Gini-importance \\
\midrule
nirDrop0 & 0.044 &    & bareSoil20 & 0.055 \\
MIRIBIMax & 0.044 &    & nirDrop1 & 0.047 \\
redDrop1 & 0.041 &    & blueMax & 0.042 \\
blueAvg & 0.041 &    & CIDrop0 & 0.039 \\
total n sentinel images & 0.041 &    & nir20 & 0.038 \\
nirDrop1 & 0.039 &    & baiDrop0 & 0.037 \\
blueMax & 0.039 &    & greenMax & 0.036 \\
MIRIBISpike0 & 0.038 &    & greenMin & 0.035 \\
redMin & 0.037 &    & nirDrop0 & 0.035 \\
red20 & 0.037 &    & nirDrop2 & 0.035 \\
NBRAvg & 0.036 &    & CI80 & 0.034 \\
SWIR2Avg & 0.036 &    & blueAvg & 0.034 \\
redEdge3Min & 0.036 &    & ndviMin & 0.033 \\
BurnScarMax & 0.036 &    & red80 & 0.032 \\
BurnScarSpike0 & 0.035 &    & bareSoilDrop1 & 0.032 \\
nirAvg & 0.033 &    & baiAvg & 0.032 \\
CIDrop0 & 0.033 &    & msaviAvg & 0.032 \\
NBRDrop0 & 0.033 &    & nirAvg & 0.031 \\
CIDrop1 & 0.033 &    & ndviAvg & 0.030 \\
CIMin & 0.032 &    & srDrop2 & 0.030 \\
nirMin & 0.032 &    & redDrop0 & 0.030 \\
SWIR1Avg & 0.032 &    & redMin & 0.029 \\
NBR2Min & 0.031 &    & srAvg & 0.029 \\
SWIR1Min & 0.031 &    & blueMin & 0.029 \\
SWIR2Min & 0.031 &    & CISpike1 & 0.028 \\
NBR2Drop1 & 0.031 &    & red20 & 0.028 \\
greenMax & 0.030 &    & greenAvg & 0.027 \\
& &    & redAvg & 0.024 \\
& &    & CIMin & 0.023 \\
\bottomrule
\end{tabular}  
\end{center}
 \footnotesize{Notes: Drop0/Drop1/Drop2 is the maximum difference (Vt+1 - Vt) in the negative direction. Spike0,Spike1/Spike2 is the maximum difference (Vt+1 - Vt) in the positive direction. 0/1/2 indicates the number of subsequent images that must be above the mean for a drop/spike to be registered.}
\end{table}

\end{document}